\documentclass[10pt,twocolumn,letterpaper]{article}

\usepackage{cvpr}
\usepackage{times}
\usepackage{epsfig}
\usepackage{graphicx}
\usepackage{amsmath}
\usepackage{amssymb}

\usepackage[pagebackref=true,breaklinks=true,letterpaper=true,colorlinks,bookmarks=false]{hyperref}

\cvprfinalcopy % *** Uncomment this line for the final submission

\def\cvprPaperID{5320} % *** Enter the CVPR Paper ID here

\ifcvprfinal\fi
\begin{document}

%%%%%%%%% TITLE
\title{In Defense of the Classification Loss for Person Re-Identification}

\author{
Yao Zhai$^{1}$\thanks{This work was done during an internship at Microsoft Research Asia.}\qquad Xun Guo$^2$\qquad Yan Lu$^2$\qquad Houqiang Li$^1$\\
\\$^1$University of Science and Technology of China\qquad $^2$Microsoft Research Asia\\
{\tt\small zy92918@mail.ustc.edu.cn\quad \{xunguo, yanlu\}@microsoft.com\quad lihq@ustc.edu.cn}
}
\maketitle
\thispagestyle{empty}

\begin{abstract}
  The recent research for person re-identification has been focused on two trends. One is learning the part-based local features to form more informative feature descriptors. The other is designing effective metric learning loss functions such as the triplet loss family. We argue that learning global features with classification loss could achieve the same goal, even with some simple and cost-effective architecture design. In this paper, we first explain why the person re-id framework with standard classification loss usually has inferior performance compared to metric learning. Based on that, we further propose a person re-id framework featured by channel grouping and multi-branch strategy, which divides global features into multiple channel groups and learns the discriminative channel group features by multi-branch classification layers. The extensive experiments show that our framework outperforms prior state-of-the-arts in terms of both accuracy and inference speed.
\end{abstract}

\section{Introduction}

Person re-identification (re-id), targeting at probing person from a large gallery set, is attracting more and more attention for its importance in video surveillance applications. With the rapid advancement of deep learning, ConvNets~\cite{alexnet,vgg,googlenet,resnet} well designed for image classification tasks~\cite{imagenet} have also realized impressive representations of person image features in person re-id, outperforming traditional handcrafted low-level features by a large margin~\cite{pcbr40}. However, person re-id remains a very challenging task, especially in real deployment, due to the dramatic variations in illumination, human body poses, camera viewpoints, background clutter, and occlusions. Since there is no overlap between training and testing categories (i.e., person identities), the task requires more discriminative feature representations to distinguish unseen similar images.

To form a better description of a person's visual appearance, there has been a research trend that aims at developing more discriminative and robust feature representations.  These methods usually generate several human body parts first, and then evolve these parts into feature learning to obtain discriminative part feature representations. Recently, Zhang et al.~\cite{alr} design a shortest path loss for aligned local parts. By jointly learning global features and local features, it achieves better person re-id performance. It is worth mentioning that in the inference stage, only using the global features is almost as good as the combined features. This indicates that the potential of global feature has not been fully exploited.

\begin{figure}[t]
\includegraphics[width = 0.5\textwidth]{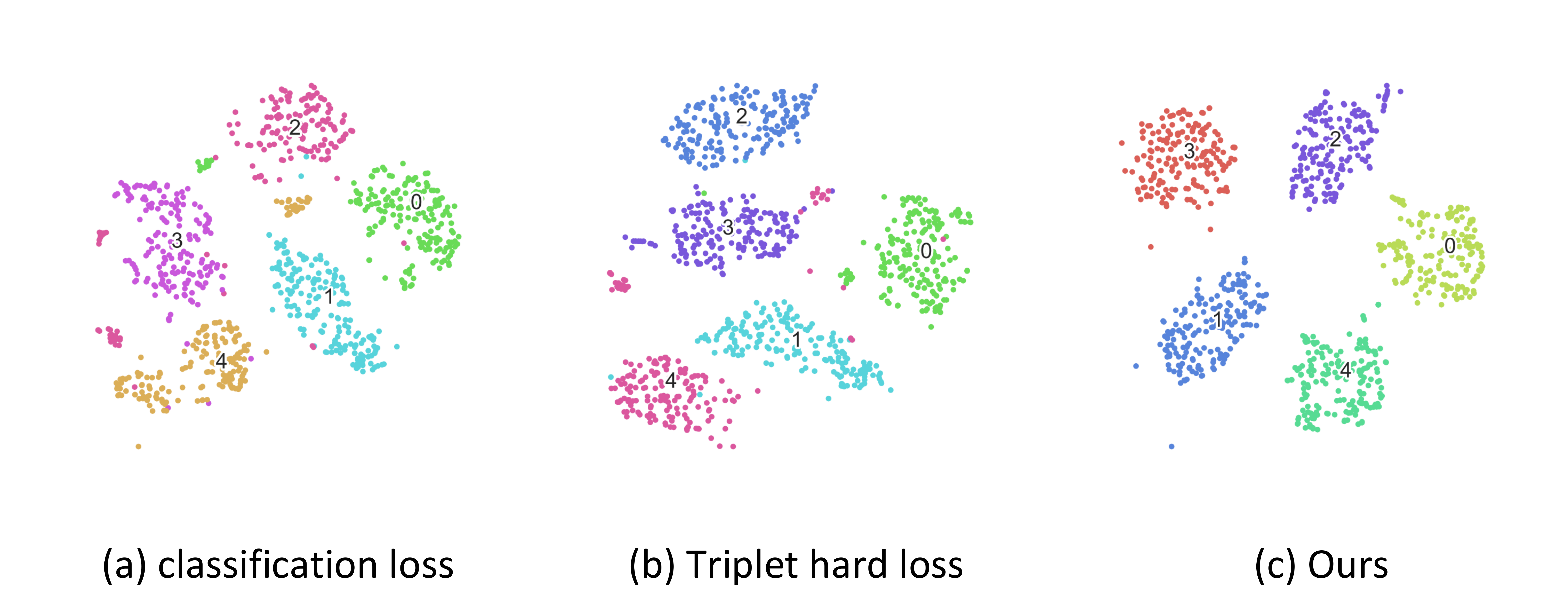}
\caption{Visualization of person features learned by (a) standard classification loss, (b) triplet hard loss, (c) our multi-branch classification loss. We choose five different testing identities to visualize their features (Best viewed in color)}
\label{fig:1}
\end{figure}

The effectiveness of dividing human body into parts has been verified. The learned local features are associated with parts of the body for visual appearance. Is it possible to partition global feature into multiple channel groups and learn channel group features?
Global feature partition is more robust than human body part partition, which suffers from misalignments caused by inaccurate bounding box detection, human pose changes and various human spatial distributions.
At the same time, different channels of the global features also have different recognition patterns. For example, some channels may focus on color information and some others may prefer texture information. Therefore, different channels of global features may also pay attention to different parts of human body. Global feature also implicits detailed part information of person bodies.

Aside from the research on learning part-based features, many existing studies are working on designing better metric learning loss functions~\cite{alr18,impTrip,triplethard,quad,alr32,hap}, including triplet loss, triplet hard loss, quadruplet loss, HAP2S loss, etc. These studies try to improve the training model's generalization capability by reducing the intra-class variations and enlarging the inter-class variations. It is worth mentioning that the performance of metric loss is significantly influenced  by  the  sampling  method. The key of utilizing metric loss is to design the hard sample mining techniques. However, it isn't always robust especially when there are outliers in the training set. Different from metric learning methods, there are approaches which address the person re-id problem from the classification aspect. These works compute the cross-entropy softmax classification loss for person identities or pair-wise images. During inference, the classification based techniques also need to compute the distance matrix of features to distinguish unseen images.

In common sense, the metric learning loss performs better than classification loss in person re-id tasks. We argue that the inferior performance of classification loss is due to the mismatch between training target and testing target. To remove this mismatch, we design a simple yet effective architecture: multi-branch fully-connected (fc) layers to learn more robust person features.

We further propose a cost-effective person re-id framework based on channel group learning. Our framework divides global feature into multiple channel groups, and learns transformed discriminative features of each group by a convolutional layer and multiple classification layers. During inference, we can use one of the channel groups or concatenate all channel groups together to formulate the feature descriptor for each person image.

Figure~\ref{fig:1} shows the visualizations of the learned features from different loss functions. Our multi-branch based classification loss achieves larger inter-class variances and smaller intra-class variances than others. The effectiveness of our method is also shown through experiments on Market-1501~\cite{pcbr39}, DukeMTMC-reID~\cite{pcbr26,pcbr43} and CUHK03~\cite{pcbr18} benchmarks. Our framework based on {\bf global features} and {\bf classification loss} achieves a state-of-the-art performance with a much smaller inference cost compared to the prevalent part-based re-id frameworks. Our contributions can be summarized into two folds:
\begin{itemize}
  \item We propose a simple yet effective design with {\bf multi-branch} classification layers for re-identification tasks. The multi-branch fc layers enable the classification loss to learn more robust features and achieve better performance compare to the standard one fc architecture.
  \item Our framework learns global feature with {\bf channel grouping}, which further achieves better performance than prevalent part-based models. At the same time, our channel grouping design has a much smaller inference cost, which is very valuable for real-time applications.
\end{itemize}

\section{Related Work}
\noindent {\bf Hand-crafted Feature Partition.} Before deep neural networks spring up, there are a great amount of research efforts~\cite{pcbr8,pcbr41,pcbr24,pcbr20,pcbr11,pcbr4,pcbr6} for designing robust handcraft partitioned features, such as color histograms, local binary patterns, Gabor features, etc. These works are effective for mitigating the variations in lighting, poses and viewpoints. The partition schemes includes horizontal stripes, body parts and patches, which are all performed in the spatial dimension.

\noindent {\bf Distance Metric and Classification Loss.} Different from hand-crafted feature design, deep neural networks attempt to automatically learn features and metrics from large constructed datasets. A group of research view person re-id as a ranking issue. Ding et al.~\cite{quar7} use a triplet loss to compute the relative distance between images. Hermans et al.~\cite{triplethard} improve the performance of Triplet loss by designing hard example mining strategies. Chen et al.~\cite{quad} introduce a quadruplet loss which enlarges inter-class variations and reduces intra-class variations. Yu et al.~\cite{hap} propose a soft hard-sample mining scheme by adaptively assigning weights to hard samples.

Meanwhile, there are approaches which address the person re-id problem from the classification aspect. Some of them compute the cross-entropy loss for image pairs in their networks~\cite{pcbr18,quar1,quar35}. Their networks take pair-wise images as inputs, and output the verification probability. Some others design a margin-based loss~\cite{quar32,quar31} to keep the largest separation between positive and negative pairs. Besides, some methods~\cite{pcbr19,pcbr35,pcb,multiloss} adopt the simple classification network performed on multiple local parts of a single image. In this paper, we adopt multiple softmax losses applied on multiple channel groups on the same image, which shows a supreme performance with much smaller training and inference cost.

\noindent {\bf Deep Learning for Local Parts.} Recently, many works learn deep representation of local parts for a more discriminative feature representation. Some works directly divide images into local stripes, which suffer from inaccurate part localization. Thus, several recent works~\cite{pcbr38,pcbr27,glad,alr50} try to align local parts by pose estimation and region proposal generation. Zhao et al.~\cite{pcbr37} design a part-aligned network for better body parts partition. Zhang et al.~\cite{alr} partition the body parts into horizontal stripes and compute the local path loss along with the global loss. Sun et al.~\cite{pcb} perform a uniform partition strategy and divide person image into horizontal stripes. By multiple classification supervision on each horizontal stripe, they achieve state-of-the-art performance.

Unlike these works that exploit spatial information, we firstly perform the partition strategy on the channel dimension of the global features. By learning multiple channel group features, our network achieves state-of-the-art re-id accuracy, which is comparable to the performance of part partition solutions. Moreover, the channel partition strategy has an additional gain in collaboration with part partition strategy, setting the new state of the art. Meanwhile, the channel group learning framework leads to a much smaller inference cost, which is practical for real-time person re-id applications. These results show the effectiveness of channel-wise feature partition.

\begin{figure}[t]
\includegraphics[width = 0.5\textwidth]{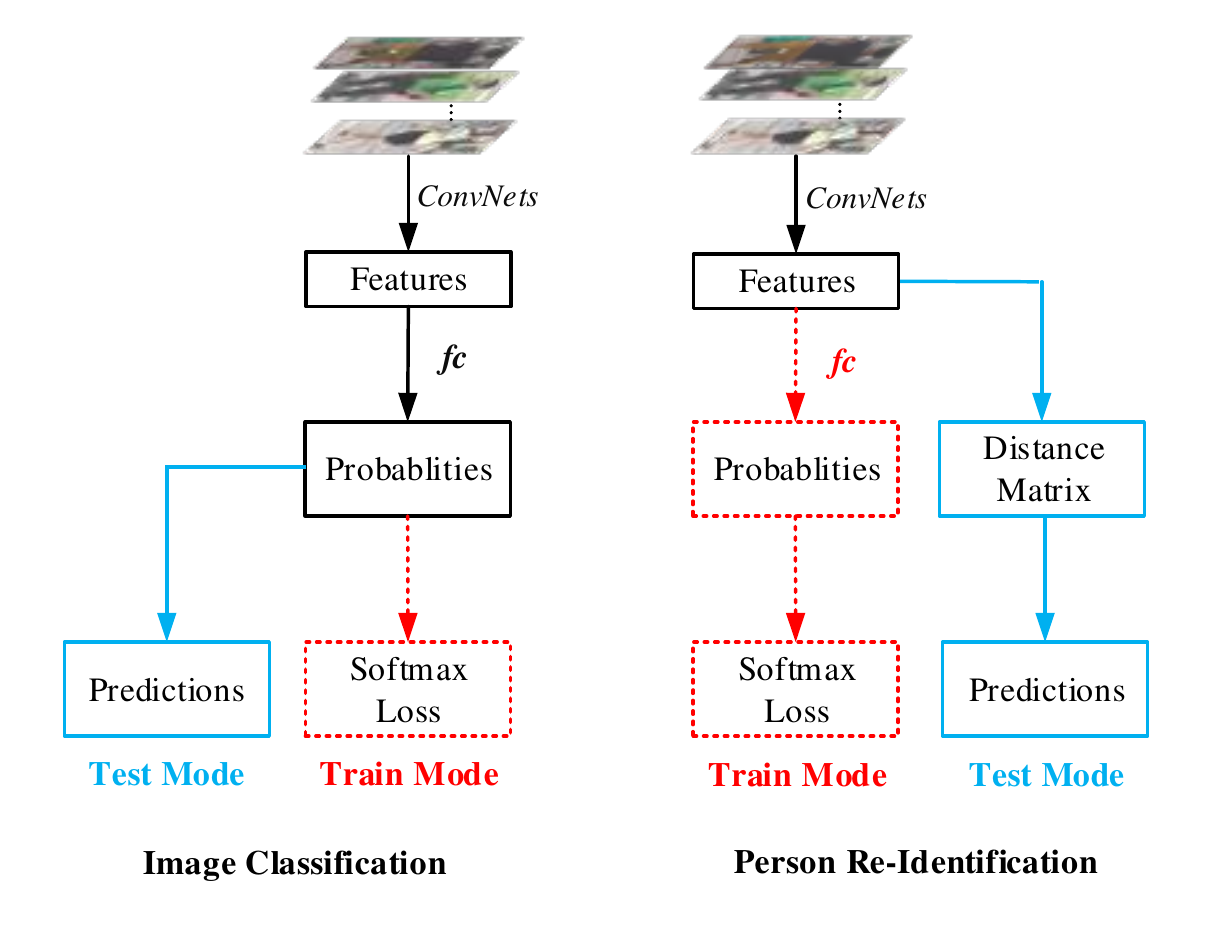}
\caption{Training and testing frameworks of image classification and person re-identification driven by cross-entropy classification loss. The most important classification (fc) layer for training is {\bf useless} during inference for person re-identification.}
\label{fig:2}
\end{figure}

\section{Problem Analysis}

\subsection{Problem Definition}

Given a set of $n$ training images $\{I_{n_i}\}_{n_i=1,...,n}$ which contain the visual appearance of $n_{id}$ different people, with
the corresponding identity labels $\{y_{n_i}\}_{n_i=1,...,n}$, $y_{n_i} \in [1, ..., n_{id}]$, our target is to identify a probe image from the unseen test gallery set. Adopting cross-entropy classification loss, the training process in person re-id is the same as the general image classification task, which pays more attention on learning better parameters for the last classification (fc) layer that serves to better predict the person identity labels on the training set. During testing process, person re-id task requires to compute distance matrix that is totally depended on the global feature vector since there is no overlap between training and testing sets.

\subsection{The drawbacks of Metric Loss}

Deep metric learning(\eg Triplet Loss) provides an effective methodology for person re-identification task. The training target of metric loss is to force the distance between intra-class triplets less than the distance between inter-class ones by at least a margin. In practice, most of the triplets already satisfy the optimization goal, thus are useless for training. Therefore, hard-sample mining is crucial in metric loss design. Generally, the hard-sample mining based metric learning methods have several drawbacks:

\noindent {\bf Data utilization.} Metric loss does not fully utilize the training set. It only verifies the labels of two samples, but ignores the specific class ID. Metric loss requires the training mini-batches to be constructed by positive samples and negative samples. Since there are much more negative samples, most of them are ignored during training. For hard sample mining based methods, most of the triplets are abandoned, which can also contribute to the network optimization.

\noindent {\bf Model robustness.} Metric loss is easily influenced by a few error-labeled samples. Since the metric loss highly relies on hard sample mining and the error-labeled samples usually act as the hardest samples, the model is probably optimized to the direction of overfitting the outliers. The robustness of models trained by metric loss will be worse if there are more bad annotations in the training set.

\subsection{In Defense of Classification Loss}

The classification loss for re-id is designed for image classification task and directly uses the multi-class labels as the supervision information. The target of softmax loss is classifying the features into predefined categories, which is totally different from that of the metric loss. The transformations contained in the classification (fc) layer play the key role to convert the decision boundary from the probability space to the feature space. We argue that the inferior performance of classification loss for person re-id task is due to the mismatch between training and testing process.

Figure~\ref{fig:2} shows the training and testing frameworks of image classification and person re-identification driven by the standard cross-entropy classification loss. The key difference between image classification and person re-identification is the inference process. The training target of the classification loss is to perform better prediction on the person identity labels, which depends more on the classification (fc) layer that is abandoned during inference for person re-id. The testing target is to compute robust distance matrix between unseen person images, which totally depends on the feature representation. To obtain a more robust person re-id model driven by classification loss, we need to increase the impact of feature learning and reduce the overfitting risk from the classification layer during training process.

\begin{figure}[t]
\includegraphics[width = 0.5\textwidth]{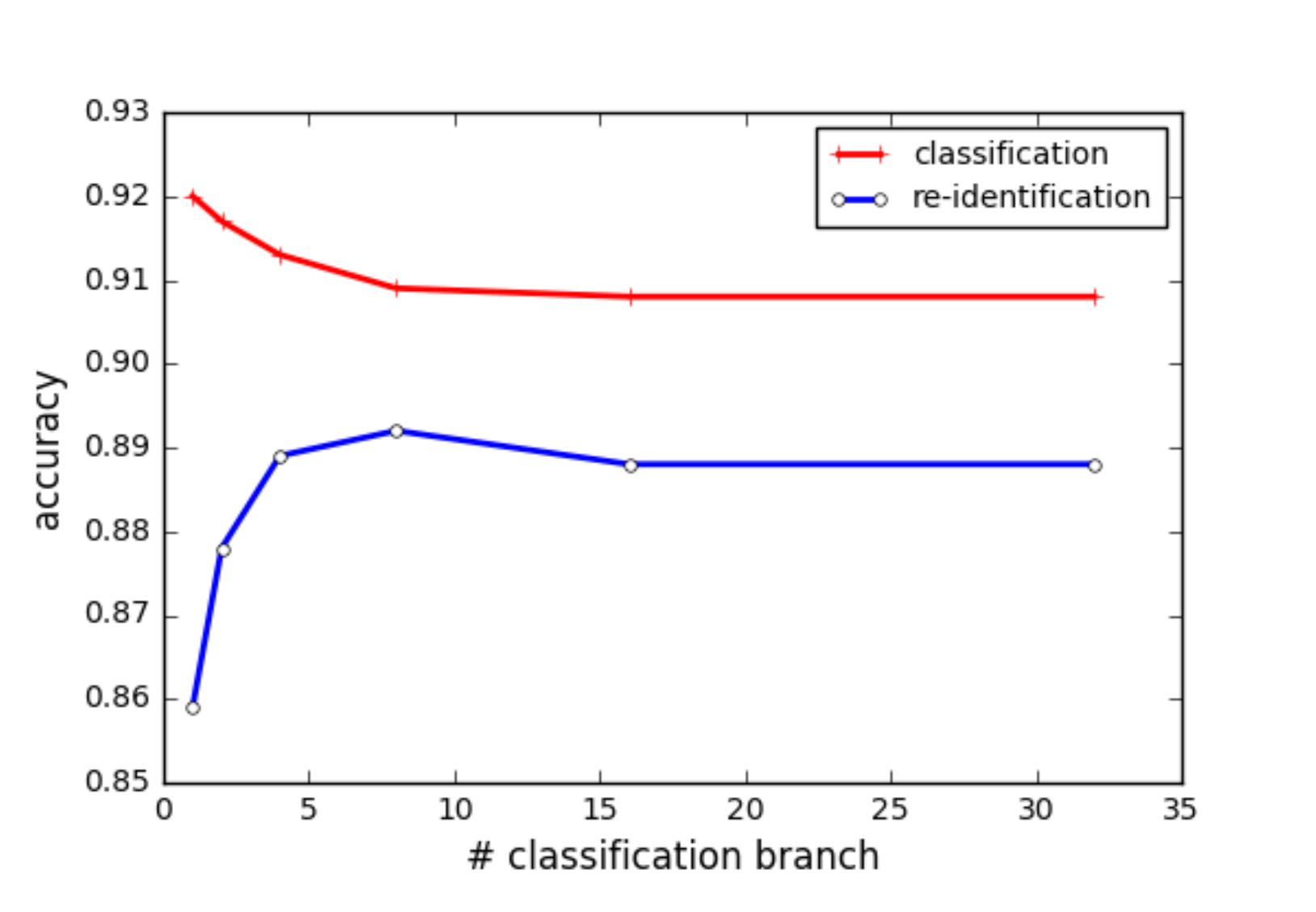}
\caption{Image classification and person re-id accuracy vs. classification branch number. We evaluate the model performance with ResNet-50 as backbone and multiple fc layers as classifiers on Market-1501 dataset. }
\label{fig:3}
\end{figure}

\subsection{Multi-branch classification layers}

To solve this problem, we propose a simple yet effective design: multi-branch classification layers. During backpropagation, the gradients from multiple fc layers gather into previous convolutional layers and make the learned classification model emphasize more on the global feature vector for computing distance matrix.

To verify this idea, we design experiments on Market-1501 dataset. We change the original one fc classifier to multi-branch fc layers (architecture (d) in Figure~\ref{fig:5}) and test the classification accuracy on validation set and the re-id accuracy on test set. As shown in Figure~\ref{fig:3}, the multi-branch classification layers cannot improve the classification accuracy on known categories, but greatly improve the re-id performance on unseen testing images. The use of multi-branch classification layers alone can bring a 3\% increase of re-id accuracy, compared to the standard one-fc classification architecture (architecture (b) in Figure~\ref{fig:5}).

\section{Our Approach}

Equipped with the multi-branch classification layers, we formulate a channel group learning CNN model. In contrast to most existing re-id methods that typically depend on exploiting part-based local features, our channel group learning framework aims to discover and capture concurrently discriminative information about a person image from different channel groups of the global feature. For each channel group, we add the same classification supervision.

\begin{figure*}[t]
\includegraphics[width = 1.0\textwidth]{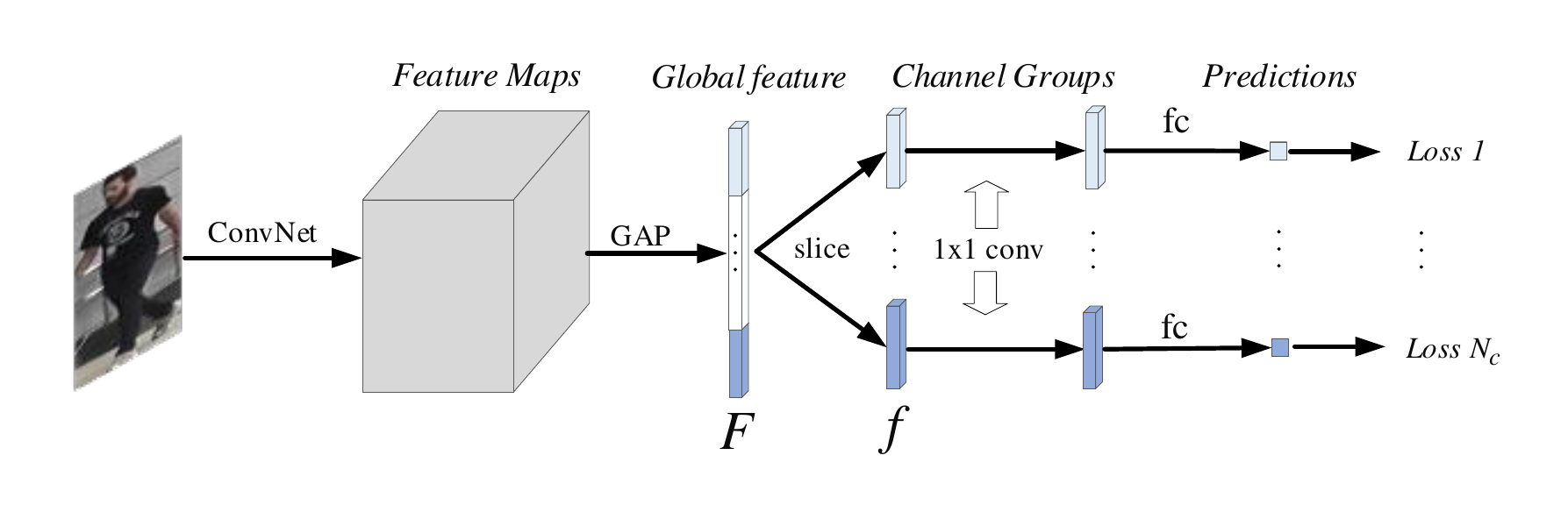}
\caption{General architecture of our method. (1) Generate the convolutional feature maps of the input person image by ConvNet; (2) perform global average pooling on the feature maps after the last convolutional layer and output the global feature $F$; (3) divide the global feature into $N_c$ channel groups $f$; (4) for each channel group, generate the transformed channel group features by a shared $1\times1$ convolutional layer; and (5) for each channel group, predict the identity by its own fc layer driven by the cross entropy classification loss.}
\label{fig:4}
\end{figure*}

\subsection{General Architecture}

The general architecture is depicted in Figure~\ref{fig:4}. Our network firstly forwards an input image through ConvNet and produces its convolutional feature maps. Then we apply a global average pooling (GAP) layer after the last convolutional layer of feature maps and get the global feature vector $F$.

After that, we perform a series of slice operations to separate the global feature into $N_c$ channel groups. Each channel group represents for partial global characteristics of the input person image. $N_c$ $1\times1$ convolutional layers are applied to the channel groups, outputting the transformed group features. The parameters of $N_c$ convolutional layers are shared between channel groups. Finally, we generate the identity predictions for each person image by $N_c$ corresponding fully-connected layers. The identity predictions are fed into following cross-entropy softmax loss layers to compute the classification loss for each channel group.

\subsection{Design Details}

\noindent {\bf Shared Convolutional layers.}
The added $1\times1$ convolutional layers are responsible for learning the most discriminative global level features from the entire person image. To concurrently optimize the feature representations for each channel group and discover the correlated information between all channel groups, the added convolutional layers are designed with parameter sharing. Sharing parameters can give each channel group more constraints during training process. Since the original channel groups sliced from global features have different recognition patterns, utilizing convolutional layers of the same parameters can make each transformed channel group keep different recognition patterns from the others.

\noindent {\bf Channel Group Division.}
We perform uniform partition on the global feature to obtain different channel groups. Given that the global feature has total $C$ channels, we separate it into $N_c$ channel groups and each channel group has $C_{g} = C/N_c$ channels. The feature value of $c$-th channel for the $i$-th channel group is obtained by,
\begin{equation}
 f_{i}(c) = F(c+(i-1)C_{g})
\end{equation}
where $ i=1,...,N_c$ and $c=1,...,C_{g}$.
Thus, there is no channel overlap between any two channel groups. We have also tried different division options of overlapped channel groups. The performance of overlapped division is inferior to the non-overlapped one.

\noindent {\bf Loss Function.}
We utilize the cross-entropy classification loss function to optimize each channel group branch given training labels of multiple person classes.
For $f_{i}$, the $i$-th channel group, the classification loss $L_{i}$ is computed by,
\begin{equation}
L_{i} = -\sum_{k=1}^{n_{id}}1\{y_{n_i} = k\} \log\frac{e^{\hat{y}_{n_i}^{i}}}{\sum_{l=1}^{n_{id}}{e^{\hat{y}_{l}^{i}}}} \quad i=1,...,N_c
\end{equation}
where $\hat{y}_{n_i}^{i}$ is the predicted person classification score of the $n_i$-th training sample. $n_{id}$ is the total category number of person identities. The total loss is obtained by,
\begin{equation}
L = \sum_{i=1}^{N_c}L_{i}
\end{equation}

The classification loss function differs significantly from triplet loss and contrastive loss functions, which are designed to exploit pairwise re-id labels and highly rely on hard example mining.  Equipped with channel grouping and multi-branch design, the relative accuracy gain of cross-entropy classification loss is much larger than the triplet loss families. The triplet hard loss with multiple channel group learning only achieves 0.5\% improvement compared to its original version, while the classification loss gets more than 6\% boost.

\noindent {\bf Inference Setting.}
Once the network is learned, we use $L_2$ distance metric to compute the distance matrix between the feature descriptors of the query image and gallery images. Our method is flexible to offer several options for inference:

\noindent {\bf Standard.} The standard setting is concatenating all channel groups, which is actually the global feature $F$ of the input image.

\noindent {\bf Fast.} The fast setting is using one channel group as the feature descriptor, because each channel group is the global representation of person image. This setting has a much smaller inference cost than the standard setting with a marginal accuracy drop.

\noindent {\bf Voting.} After computing the distance matrices of each channel group, we get the different re-id results. We can count the re-id results and vote for the final re-id result. The voting setting gives a continual performance gain with little additional inference cost compared with the standard setting.

\begin{figure*}[t]
\includegraphics[width = 1.0\textwidth]{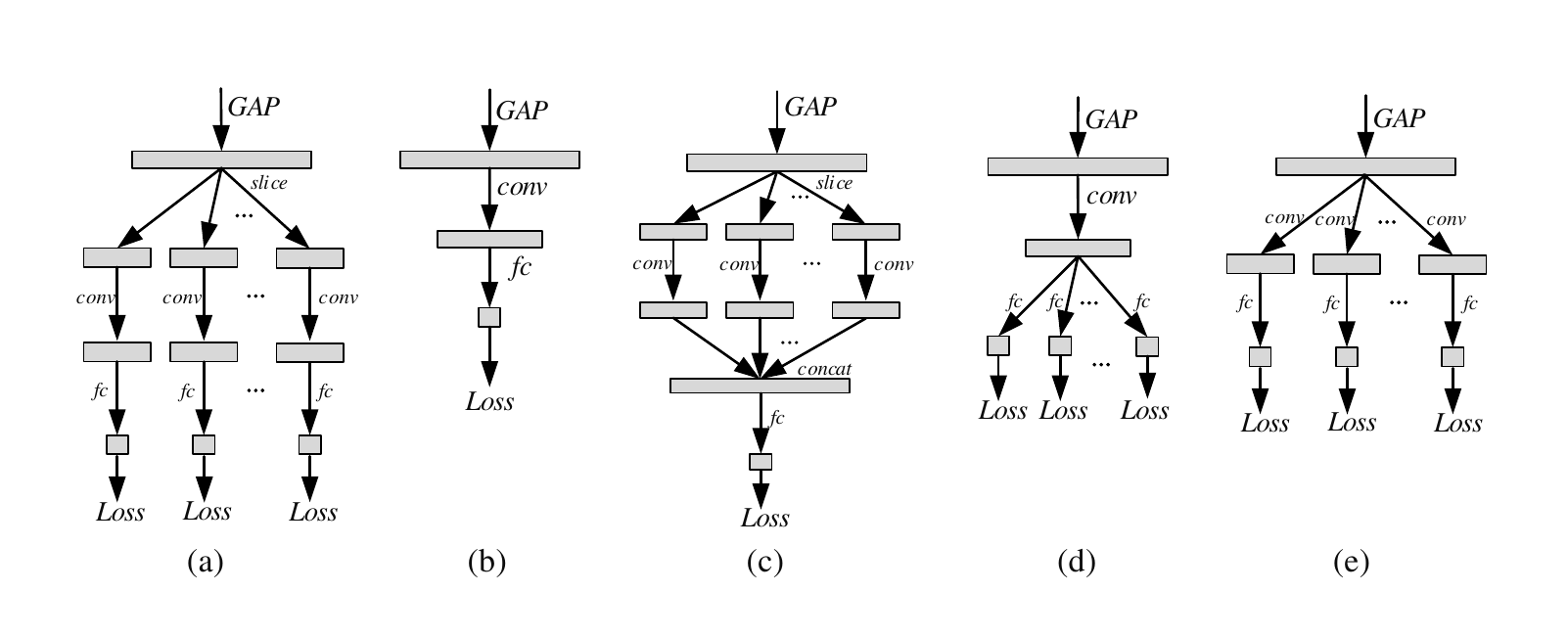}
\caption{Different architectures driven by classification loss. The inputs are convolutional feature maps. (a) Our channel group learning framework; (b) standard classification framework; (c) channel grouping without multi-branch classification layers; (d) multi-branch classification layers without channel grouping; (e) (d) with multi-branch convolutional layers.}
\label{fig:5}
\end{figure*}

\subsection{Architecture Variants}
Our channel group learning framework has two key designs: channel grouping and multiple classification layers. Figure~\ref{fig:5} shows some possible architecture variants all driven by the classification loss. Architecture (a) is our method, which divides global features into multiple channel group and then perform multi-branch strategy. (b) - (e) are the variants of our method.

\noindent{\bf Channel grouping design.} (b) is the original standard classification architecture. (c) applies the channel grouping design, but abandons the multi-branch classification layers. This architecture is equivalent to (b) with a grouped convolutional layer.

\noindent{\bf Multi-branch fc design.} (d) and (e) adopt multi-branch classification layers and compute multiple cross entropy losses. (e) uses multiple different convolutional layers for each branch. (d) and (e) don't employ the channel group design, which divides the global feature into multiple groups and then performs multi-branch operations.

\begin{table}[t]
\begin{center}
\begin{tabular}{c|cc|cc|cc}
\hline
& & & \multicolumn{2}{|c|}{Market-1501}  & \multicolumn{2}{|c}{DukeMTMC} \\
Design & {\bf CG} & {\bf MB} & Rank-1 & mAP  & Rank-1 & mAP \\
\hline
(a) & $\checkmark$ & $\checkmark$ & {\bf 92.6} & {\bf 78.3} & {\bf 82.8} & {\bf 66.7}      \\
(b)  & & & 85.9 & 63.9 & 73.5 & 53.1    \\
(c) & $\checkmark$ & & 86.6 & 64.5 & 73.7 & 53.3   \\
(d) & & $\checkmark$ & 89.2 & 76.5 & 79.7  &  59.5  \\
(e)  & & $\checkmark$ & 89.6 & 76.8 & 79.9 &  60.1  \\
\hline
\end{tabular}
\end{center}
\caption{Experiment results on Market-1501 of different architecture designs in Figure~\ref{fig:5}. {\bf CG} represents for channel group strategy, {\bf MB} represents for multi-branch strategy.}
\label{lab:2}
\end{table}

Here we perform the experiments and evaluate the re-id performance of different architectures shown in Figure~\ref{fig:5}. The channel group number $N_c$ is set to 8. Table~\ref{lab:2} shows that, compared to the standard classification architecture design (b), only using channel group strategy without multi-branch classification layers improves the baseline by 0.7\%. Grouped convolutions without multi-branch strategy could hardly achieve a large performance gain. Once equipped with multiple classification layers, the network gets 89.6\% rank-1 accuracy. With channel grouping help, we further achieve 92.6\% rank-1 accuracy on Market-1501 dataset, which is a 6.7\% gain compared to the standard classification baseline. The multi-branch fc and channel grouping strategy are both necessary for achieving state-of-the-art performance.

\section{Experiments}
In the next, we perform several experiments to evaluate the effectiveness of the proposed method under different parameter settings.

\noindent {\bf Datasets.} We evaluate the proposed method on three widely used person
re-id datasets: Market-1501~\cite{pcbr39}, DukeMTMC-reID~\cite{pcbr26,pcbr43} and CUHK03~\cite{pcbr18}.

Market-1501 is composed of 19,732 gallery images, 3,368 query images and 1,501 identities automatically detected from six cameras. The training set contains 12,936 images
of 751 identities. The testing set has 19,732 images of 750 identities.

DukeMTMC-reID dataset contains 36,411 images of 1,812 identities shot by eight high-resolution cameras. There are 16,522 images of 702 identities in the training set and 2,228 query images of the other 702 identities and 17,661 gallery images. It is one of the largest challenging pedestrian image datasets.

CUHK03 contains 17,097 images of 1,467 identities captured from two cameras. Each identity contains 9.6 images on average. It provide both detected images by pedestrian detector and
human labeled bounding boxes. It provides 20 split sets, each randomly selects 1,367 identities for training and the rest for testing. We follow the new training/testing protocol proposed in ~\cite{pcbr44}.

\vspace{6pt}
\noindent{\bf Experimental Settings.}
We choose ResNet-50 as our network backbone. Following the modification in R-FCN~\cite{rfcn} and PCB~\cite{pcb}, we change the stride of the last downsampling block from 2 to 1, which makes the spatial size of convolutional feature maps larger before global average pooling. After uniform splitting the global features into $N_c$ channel groups, we adopt $1\times1$ convolutional layers to transform the channel groups into 256-d feature vectors. Each added convolutional layer is followed by a Batch Normalization layer~\cite{batchnorm} and an ReLU layer. The mini-batch size is set to 128, in which each identity has 4 images. The training images are resized to $256\times128$ pixels. The data augmentation includes horizontal flipping and random cropping. We take the standard inference setting if not mentioning. We mainly report the performance of single query mode without using any re-ranking algorithm which considerably improves the performance especially mAP.

\subsection{Channel Group Learning}

\begin{table}[h]
\begin{center}
\begin{tabular}{c|cc|cc}
\hline
& \multicolumn{2}{|c|}{Market-1501} & \multicolumn{2}{|c}{DukeMTMC} \\
$N_c$ & Rank-1 & mAP & Rank-1 & mAP \\
\hline
1(TripHard Loss) & 88.1 & 70.5  & 76.6 & 57.7 \\
8(TripHard Loss) &  88.6 &  70.9  & 76.8 & 58.2 \\
\hline
1 & 85.9 & 63.9  & 73.5 & 53.1   \\
2 & 90.1 & 71.7  & 80.0 &  61.4  \\
4 & 92.0 & 77.5  & 83.0 & 66.5    \\
8 & {\bf 92.6} & {\bf 78.3}  & 82.8 & 66.7 \\
16 & 91.9 & 77.4  & {\bf 83.6}  & {\bf 67.4}   \\
32 & 90.4 & 74.2 & 79.0 & 59.9   \\
\hline
\end{tabular}
\end{center}
\caption{Experiment results of different channel group number $N_c$.}
\label{lab:1}
\end{table}

First of all, we investigate the performance of channel group learning under different settings of channel group number $N_c$. We change the channel group number from 1 to 32 and evaluate the performance on Market-1501 and DukeMTMC-reID.  Table~\ref{lab:1} shows the effectiveness of channel group learning. The performance of standard classification loss is inferior to the TripHard loss~\cite{triplethard}. After partitioning the global feature into two channel groups, we receive a 4.2\% gain of rank-1 accuracy and a 7.8\% gain of mAP on Market-1501. When $N_c$ is set to 8 or 16, we get the best performances on two datasets, which are more than 6\% rank-1 accuracy improvement and 14\% mAP improvement. The TripHard loss with channel group learning achieves much smaller improvement compared to classification loss. These results verify the effectiveness of channel group learning driven by classification loss.

\subsection{Sharing Convolutional Layer}

\begin{table}[h]
\begin{center}
\begin{tabular}{c|c|cc|cc}
\hline
& & \multicolumn{2}{|c|}{Market-1501}  & \multicolumn{2}{|c}{DukeMTMC} \\
$N_c$ & shared & Rank-1 & mAP  & Rank-1 & mAP \\
\hline
2 &  & 88.3 & 70.1 & 78.5 & 59.9      \\
  & $\checkmark$ & 90.1 & 71.7 & 80.0 & 61.4    \\
\hline
4 & & 90.1 & 74.9 & 81.5  &  63.8   \\
  & $\checkmark$ & 92.0 & 77.5 & 83.0 & 66.5   \\
\hline
8 &  & 90.8 & 76.5 & 81.8 & 65.8   \\
  & $\checkmark$ & 92.6 & 78.3 & 82.8 & 66.7  \\
\hline
\end{tabular}
\end{center}
\caption{Experiment results of sharing convolution parameters or not. }
\label{lab:3}
\end{table}

In our network, the added $1\times1$ convolutional layers are parameter sharing. Here we continue the experiments and evaluate the necessity of sharing parameters between the added convolutional layers. We choose three settings of channel group number: $N_c = 2, 4, 8$, and test the performance of sharing convolution parameters or not.

The results in Table~\ref{lab:3} show that, sharing parameters between added convolutional layers generally brings a 2\% accuracy improvement. This indicates that sharing convolution parameters between multiple channel groups is necessary and effective.

\subsection{Inference Settings}

\begin{table}[t]
\begin{center}
\begin{tabular}{cc|c|cc|cc}
\hline
& & & \multicolumn{2}{|c|}{Market-1501}  & \multicolumn{2}{|c}{DukeMTMC} \\
$N_c$ & $C_g$ & $Dim_f$ & Rank-1 & mAP  & Rank-1 & mAP \\
\hline
2 & 1024 & 1024 & 89.6 & 71.1 & 80.0 & 61.0  \\
  &  & 2048 & 90.1 & 71.7 & 80.0 & 61.4  \\
  &  & voting & 90.0 & 71.4 & 80.1 & 61.5  \\
\hline
4 & 512 & 512 & 91.8 & 76.4 & 82.8 &  65.5  \\
  &  & 1024 & 92.0 & 76.9 & 82.8 &  66.3  \\
  &  & 2048 & 92.0 & 77.5 & 83.0 & 66.5   \\
  &  & voting & 92.3 & 77.9 & 83.2 & 66.9   \\
\hline
8 & 256 & 256 & 91.7 & 77.1 & 82.3 & 65.5    \\
  &  & 512 & 91.8 & 77.4 & 82.6 &  66.3    \\
  &  & 2048 & 92.6 & 78.3 & 82.8 & 66.7   \\
  &  & voting & {\bf 93.1} & {\bf 78.9} & 83.9 & 68.2   \\
\hline
16 & 128 & 128 & 91.5 & 75.1 & 82.0 & 65.3     \\
  &  & 256 & 91.4 & 76.3 & 82.7 & 66.5     \\
  &  & 512 & 91.6 & 76.8  &  83.3 & 67.0     \\
  &  & 2048 & 91.9 & 77.4 & 83.6 & 67.4   \\
  &  & voting & 92.5 & 78.0 & {\bf 84.0} & {\bf 68.4}   \\
\hline
\end{tabular}
\end{center}
\caption{Experiment results of inference setting. $N_c$ is the channel group number. $C_g$ is the channel number of each channel group. $Dim_f$ represents for the size of feature descriptor during inference for computing distance. }
\label{lab:5}
\end{table}

\noindent{\bf Faster speed.} Prevalent studies for person re-id task usually focus on improving the re-id accuracy without considering inference speed. However, person re-id is widely used in real-time applications, such as tracking or searching people from a large gallery set, where fast inference speed is a vital component. The inference cost of person re-id mainly includes the time consumed by feature computation for each image and the time consumed by distance computation between image pairs.

The forward time depends on the input image size and network depth. The part-based models need a large input size to keep a considerable granularity of spatial size for feature maps, which introduces much forward time cost. The input size of our network ($256\times128$) is much smaller than counterparts deploying local part partitions, such as PCB~\cite{pcb} ($384\times128$), Aligned Re-ID~\cite{alr} ($224\times224$), etc.

The time cost of distance computation mainly relies on the size of feature descriptor. The computation of $L_2$ distance between $Dim_f$-d feature descriptors of an image pair costs $Dim_f$ multiply operations. Since each learned channel group is global for the whole image in our framework, we can choose any one of the channel groups or the concatenation of them to be the feature descriptor. Table~\ref{lab:5} shows the results of different feature descriptor settings in our framework. Our network is able to output a small size of feature descriptor for fast inference while keeping a high accuracy.

\noindent{\bf Higher performance.} Besides from saving inference cost, our multi-branch design can easily bring a continual performance gain by cost-free voting process. Once generated the re-id results of different channel groups, we count the re-id label results of all channel groups and get the final re-id decision. The voting process is cost-free compared to the distance computation process or any re-ranking algorithms. The total inference time is nearly equivalent to the standard inference setting ($Dim_f = 2048$).

\subsection{Integration with Part Partition}

\begin{table}[t]
\begin{center}
\begin{tabular}{c|cc|cc|cc}
\hline
& & & \multicolumn{2}{|c|}{Market-1501} & \multicolumn{2}{|c}{DukeMTMC}\\
method & {\bf C} & {\bf P} & Rank-1 & mAP & Rank-1 & mAP \\
\hline
PCB &  & $\checkmark$ &  92.4 & 77.3 & 81.8 & 66.1 \\
PCB+RPP & & $\checkmark$ & 93.8 & 81.6 & 83.3 & 69.2 \\
Ours & $\checkmark$  & & 93.1 & 78.9 & 83.9 & 68.2 \\
Ours+Part & $\checkmark$ & $\checkmark$ & {\bf 93.9} & {\bf 80.5} & {\bf 84.7} & {\bf 69.4} \\
\hline
\end{tabular}
\end{center}
\caption{Experiment results of channel partition in collaboration with part partition. {\bf C} represents for channel partition, {\bf P} represents for part partition.}
\label{lab:4}
\end{table}

Sun et al.~\cite{pcb} propose part-based convolutional baseline (PCB), which partitions images to horizontal stripes and achieves state-of-the-art performance. Here we compare the part partition with channel partition strategy by several experiments. To achieve the best performance for body part partition, we change the input size to $384\times128$ pixels and set the number of horizontal stripes to 6 following PCB. The channel group number is set to 8. During inference, we concatenate the learned channel group features and horizontal part features together into the feature descriptors.

Table~\ref{lab:4} shows that, our method has better performance than body part partition, thanks for the channel grouping and cost-free voting strategy. The large proportion of the performance gain achieved by PCB benefits from the multiple classification layers, not the superiority of body part partition, which only has a comparable performance with channel partition. After we concatenate the part features with channel group features, the channel partition strategy and part partition strategy collaborate each other and bring an additional gain.

\subsection{Compared with state-of-the-arts}

\begin{table}[h]
\begin{center}
\begin{tabular}{c|cccc}
\hline
Methods & Rank-1 & Rank-5  & mAP \\
\hline
BoW+kissme~\cite{pcbr39} & 44.4 & 63.9 & 20.8  \\
WARCA~\cite{pcbr16} & 45.2 & 68.1 & -   \\
KLFDA~\cite{pcbr17} & 46.5 & 71.1 & -   \\
SOMAnet~\cite{pcbr1} & 73.9 & - & 47.9    \\
SVDNet~\cite{pcbr28} & 82.3 & 92.3  & 62.1    \\
PAN~\cite{pcbr42} & 82.8 & - & 63.4   \\
Transfer~\cite{pcbr10} & 83.7 & -  & 65.5   \\
TripletHardLoss~\cite{pcbr10} & 84.9 & 94.2  & 69.1    \\
DML~\cite{pcbr36} & 87.7 & -  & 68.8   \\
MultiRegion~\cite{pcbr30} & 66.4 & 85.0  & 41.2   \\
HydraPlus~\cite{pcbr22} & 76.9 & 91.3  & -   \\
SpindleNet~\cite{alr50} & 76.9 & 91.5  & - \\
PAR~\cite{pcbr37} & 81.0 & 92.0  & 63.4   \\
MultiLoss~\cite{pcbr19} & 83.9 & -  & 64.4    \\
PDC~\cite{pcbr27} & 84.4 & 92.7  & 63.4    \\
PartLoss~\cite{pcbr35} & 88.2 & -  & 94.3   \\
MultiScale~\cite{pcbr3} & 88.9 & -  & 73.1    \\
GLAD~\cite{glad} & 89.9 & -  & 73.9    \\
AlignedReID~\cite{alr} & 91.8 & 97.1  & 79.3 \\
PCB~\cite{pcb} & 92.3 & 97.2  & 77.4   \\
PCB+RPP~\cite{pcb} & 93.8 & 97.5  & {\bf 81.6}   \\
\hline
Ours & 92.6 & 97.4 & 78.3   \\
Ours(with Part) & {\bf 93.9} & {\bf 97.8} & 80.5   \\
\hline
\end{tabular}
\end{center}
\caption{Compared with state-of-the-art on Market-1501.}
\label{lab:6}
\end{table}

\setlength{\tabcolsep}{6pt}
\begin{table}[t]
\begin{center}
\begin{tabular}{c|cc|cc}
\hline
& \multicolumn{2}{|c|}{DukeMTMC}  & \multicolumn{2}{|c}{CUHK-03} \\
Methods & Rank-1 & mAP & Rank-1 & mAP \\
\hline
BoW+kissme~\cite{pcbr39} & 25.1 & 12.2 & 6.4 & 6.4  \\
LOMO+XQDA~\cite{pcbr20} & 30.8 & 17.0 & 12.8 & 11.5   \\
GAN~\cite{pcbr43} & 67.7 & 47.1 & - & -   \\
PAN~\cite{pcbr42} & 71.6 & 51.5 & 36.3 & 34.0    \\
SVDNet~\cite{pcbr28} & 76.7 & 56.8 & 41.5 & 37.3   \\
MultiScale~\cite{pcbr3} & 79.2 & 60.6 & 40.7 & 37.0   \\
TriNet+Era~\cite{pcbr45} & 73.0 & 56.6 & 55.5 & 50.7   \\
SVDNet+Era~\cite{pcbr45} & 79.3 & 62.4 & 48.7 & 43.5   \\
PCB~\cite{pcb} & 81.8 & 66.1 & 61.3 & 54.2   \\
PCB+RPP~\cite{pcb} & 83.3 & 69.2 & {\bf 62.8} & {\bf 56.7}   \\
\hline
Ours & 83.9 & 68.2 & 59.9 & 53.3   \\
Ours(with part) & {\bf 84.7} & {\bf 69.4} & 61.7 & 55.3   \\
\hline
\end{tabular}
\end{center}
\caption{Compared with state-of-the-art on DukeMTMC-reID and CUHK-03.}
\label{lab:7}
\end{table}

Table~\ref{lab:6} and Table~\ref{lab:7} show the state-of-the-art results on Market-1501, DukeMTMC-reID and CUHK-03. Our channel group learning network surpasses most of the part-based models. With the help of local part partition, we further achieve a better performance and set the new state-of-the-art.

\section{Conclusion}

We have proposed a simple yet effective channel group learning framework for person re-identification based on global features and classification loss. Our framework divides global feature into multiple channel groups. With a shared convolutional layer and multiple classification layers, our network learns multiple discriminative channel group features. The simple but effective multi-branch design empower the classification loss to perform better on person re-id. The channel group features form a more robust feature representation of person images and achieve state-of-the-art performance on different person re-id benchmarks while keeping a fast inference speed. We demonstrate the channel grouping and multi-branch strategy for re-identification task. We hope these strategies could inspire researches on re-identification or other vision tasks.

{\small
\bibliographystyle{ieee}
\bibliography{ref}
}
\end{document}